\documentclass{article}
\usepackage[utf8]{inputenc}
\usepackage[T1]{fontenc}
\usepackage{graphicx}
\usepackage{grffile}
\usepackage{longtable}
\usepackage{wrapfig}
\usepackage{rotating}
\usepackage[normalem]{ulem}
\usepackage{amsmath}
\usepackage{textcomp}
\usepackage{amssymb}
\usepackage{capt-of}
\usepackage{hyperref}
\author{J.-M. Chauvet}
\date{\today}
\title{The 30-Year Cycle In The AI Debate}
\hypersetup{
 pdfauthor={J.-M. Chauvet},
 pdftitle={The 30-Year Cycle In The AI Debate},
 pdfkeywords={},
 pdfsubject={},
 pdfcreator={Emacs 25.1.1 (Org mode 9.1.13)}, 
 pdflang={English}}
\begin{document}

\maketitle
\begin{abstract}
In the last couple of years, the rise of Artificial Intelligence and the successes of academic breakthroughs in the field have been inescapable. Vast sums of money have been thrown at AI start-ups. Many existing tech companies–-including the giants like Google, Amazon, Facebook, and Microsoft--have opened new research labs. The rapid changes in these everyday work and entertainment tools have fueled a rising interest in the underlying technology itself; journalists write about AI tirelessly, and companies—-of tech nature or not--brand themselves with AI, Machine Learning or Deep Learning whenever they get a chance. Confronting squarely this media coverage, several analysts are starting to voice concerns about over-interpretation of AI's blazing successes and the sometimes poor public reporting on the topic.
This paper reviews briefly the track-record in AI and Machine Learning and finds this pattern of early dramatic successes, followed by philosophical critique and unexpected difficulties, if not downright stagnation, returning almost to the clock in 30-year cycles since 1958.
\end{abstract}
\section{Introduction}
\label{sec:orgf569a18}
The recent practical successes \cite{Hinton2012} of Artificial Intelligence (AI) programs of the Reinforcement Learning and Deep Learning varieties in game playing, natural language processing and image classification, are now calling attention to the envisioned pitfalls of their hypothetical extension to wider domains of human behavior. Several voices from the industry and academia are now routinely raising concerns over the advances  \cite{Hassabis2017} of often heavily media-covered representatives of this new generation of programs such as \emph{Deep Blue}, \emph{Watson}, \emph{Google Translate}, \emph{AlphaGo} and \emph{AlphaZero}.

Most of these cutting-edge algorithms generally fall under the class of supervised learning, a branch of the still evolving taxonomy of Machine Learning techniques in AI research. In most cases the implementation choice is artificial neural networks software, the workhorse of the Connectionism school of thought in both AI and Cognitive Psychology.

Confronting the current wave of connectionist architectures, critics usually raise issues of \emph{interpretability} (Can the remarkable predictive capabilities be trusted in real-life tasks? Are these capabilities transferable to unfamiliar situations or to different tasks altogether? How informative are the results about the real world; about human cognition? Can we infer sound causal relationships and natural explanations from them?). Others evoke higher concerns that still reverberate in Cognitive Psychology and in the Philosophy of Mind: neural plausibility of the current game-adept generation of connectionism, proper accounting for minimum innate preconditions in perception in artificial and human subjects, psychological value of laws about local, incremental connections of sub-conceptual units -- at work in artificial neurons -- versus laws about general, complex properties of large-scale representations -- found in concepts, categories, and logical rules.

Recently, ethical and moral questions were also addressed at the AI research community in the context of practical public applications of the current crop of results and their impact on society (unemployment, inequality, humanity, error correction, bias, security, unintended consequences, singularity, robot rights \cite{bostrom_yudkowsky_2014}).

This paper offers a tentative review of the roots of the growing controversy over the interpretation of AI research results, in the two previous occurrences of this same discussion, only 30 years apart. Perusing the well-documented approach \cite{Star1989,Collins1983} of elevating the status of the AI artifacts produced by research groups to one of \emph{boundary object} with other disciplines, we argue that the current wave of critical analyses re-enacts what looks like a 30-year cyclic pattern in the discussion of AI.

\section{Pattern}
\label{sec:org1c8da63}
\subsection{The Advent of Modern Computers: Connectionism and Electronic Brains}
\label{sec:orge440ec6}
In a Cornell Aeronautical Laboratory Technical Report dated January 1957 \cite{Rosenblatt1957}, Frank Rosenblatt (1928-1971) suggested that a brain model, based on his independent research on physiological psychology, should be built, not only to demonstrate the workability of the \emph{Perceptron}, as it was named, but as a research tool for further study of the principles it employed:
\begin{itemize}
\item dependence on probabilities rather than on deterministic principles for its operations, and
\item reliability gained from the properties of statistical measurements obtained on large populations of elements.
\end{itemize}
The external description of the Perceptron borrows heavily from behaviorism, which at the time was prevalent in empirical psychology, with its characteristic explanatory tools of input-input and input-output associations (e.g. in classical conditioning), and the reward-punishment track record shaping behavior (e.g. operant conditioning). In its ``photoperceptron'' instantiation, the electronic or electromechanical device performance is described as a process of learning to give the same output signal for all optical stimuli belonging to some arbitrarily constituted class. Rosenblatt summarized experimental results:

\begin{quote}
``Simple cognitive sets, selective recall, and spontaneous recognition of the classes present in a given environment are possible.''
\end{quote}

Based on these successes, ambitious goals were claimed for the project PARA (Perceiving And Recognizing Automaton):
\begin{quote}
``[\ldots{}] with the objective of designing, fabricating, and evaluating an electronic brain model, the photoperceptron. The proposed pilot model will be capable of `learning' responses to ordinary visual patterns or forms. The system will employ a new theory of memory storage (the theory of statistical separability), which permits the recognition of complex patterns with an efficiency far greater than that attainable by existing computers. Devices of this sort are expected ultimately to be capable of concept formation, language translation, collation of military intelligence, and the solution of problems through inductive logic.''
\end{quote}
and the initial results impressive and broadly publicized in the scientific (and defense) community.

\subsection{The Advent of Modern Computers: Thought and Mechanical Information Processing}
\label{sec:orge4e5ad0}
On the same year, 1957, Allen Newell (1927-1992), John C. Shaw (1922-1991) and Herbert A. Simon (1916-2001) commented on empirical explorations with the modern computer in their \emph{Logic Theorist} (LT). They sought a proper definition of heuristics as a basis for a theory of general problem solving. LT is a computer programmed to the specifications of a system for finding proofs of theorems in elementary logic, a step in an articulated research program on complex information-processing systems. This program culminated in 1961 with the seminal paper by Allen Newell and Herbert Simon introducing \emph{GPS} (General Problem Solver) \cite{Newell1958} a ``program that simulates human thought''.

Logic Theorist soon proved 38 of the first 52 theorems in chapter 2 of the \emph{Principia Mathematica}. The proof of theorem \emph{2.85} was actually advocated as more elegant than the proof produced laboriously by hand by Russell and Whitehead. GPS solved problems in symbolic logic and in trigonometry.

The experimental results led the authors to posit the GPS specification, abstracted from the symbolic logic task at hand to any task environment, as a theory of human problem solving. Simon's enthusiastic claims summed it up:
\begin{quote}
``Subsequent work has tended [\ldots{}] to demonstrate that heuristics, or rules of thumb, form the integral core of human problem-solving processes. As we begin to understand the nature of the heuristics that people use in thinking, the mystery begins to dissolve from such  vaguely understood processes as `intuition' and `judgment'.''
\end{quote}

As a theory GPS is successful at explaining the human subject problem-solving behavior, as manifested in the almost perfect match of trace and protocol, relying on  a model of symbol manipulations the complexity of which became addressable thanks recent advances made in artificial intelligence, programming, and modern computers at the time. And to quote the authors' conclusion: ``although we know this only for small fragment of behavior, the depth of the explanation is striking''.

\subsection{The Philosophical Critique (1958)}
\label{sec:org29a83ef}
Not so striking though to the eyes of the likes of philosophers Jacques Ellul (1912-1994) \cite{Ellul} and specially Hubert L. Dreyfus (1929-2017) who engaged right from these early sixties into decades-long systematic debunking of the fanciest claims by the originators of the field of AI.

In a widely disseminated article, aptly titled \emph{Alchemy and Artificial Intelligence} \cite{Dreyfus1965}, Dreyfus took stock of the then current state of the art and identified a pattern that proved recurrent:
\begin{quote}
``The field of artificial intelligence exhibits a recurrent pattern: early, dramatic success followed by sudden unexpected difficulties.''
\end{quote}
Dreyfus already read definite signs of stagnation in such typical AI research domains as language translation, game playing, pattern recognition, and problem solving. These negative results are in fact significant argues Dreyfus and should be given full consideration if progresses are to be made in the field. In Dreyfus's analysis, a system that could equal human performance at recognizing patterns, far from being readily available in 1958 as prematurely announced e.g. by Oliver G. Selfridge (1926-2008) in his landmark paper, \emph{Pandemonium: A Paradigm For Learning} \cite{Selfridge1955}, should be able to:
\begin{itemize}
\item distinguish the essential from the inessential features of a particular instance of a pattern;
\item use cues that remain on the fringes of consciousness;
\item take account of the context; and
\item perceive the individual as typical (as opposed to generic).
\end{itemize}
Furthermore, admonished Dreyfus, serious misconceptions masked the seriousness of these required abilities. Namely the assumption of ``workers in cognitive simulation'' that human and mechanical information processing involve the same elementary process. 

So does Dreyfus's stern 1958 conclusion that ``confidence of progress in cognitive simulation is thus as unfounded as the associationist assumption'' leaves a silver lining, that ``this realization, however, leaves untouched the weaker claim [\ldots{}] that human intelligent behavior -- not human information processing -- can be simulated by using digital computers''? Not even, says Dreyfus who raised a computing power argument: ``We do not know the equations of the physical processes in the brain, and even if we did, the solutions of the equations describing the simplest reaction would take a prohibitive amount of time''.

Instead, Dreyfus defined a role for AI research if directed at areas of intelligent activity categorized at level 3 on a 1-to-4 scale of ``associationistic'', ``non-formal'', ``simple formal'', and ``complex formal''. Not that effort in areas 2 (problems where insight and perceptive guesses are required, translation of natural languages, recognition of varied and distorted patterns) and 4 (incomputable games -- chess and go, mentioned -- and complex combinatory problems, or recognition of complex patterns in noise) would be wasted but would only bear fruition when, following a remark of Claude E. Shannon (1916-2001):

\begin{quote}
``efficient machines for such problem as pattern recognition, language translation, and so on, may require a different type of computer than any we have today. It is my feeling that this computer will be so organized that single components do not carry out simple, easily described functions'' 
\end{quote}

would become available. So that the critical remarks of Dreyfus can be understood as extending also to the contemporary Rosenblatt's photoperceptron as a step towards an electronic brain.

The actual death spell to Rosenblatt's perceptron intuition, inspired by the works of Donald O. Hebb (1904-1985) in 1949, and of Warren S. McCulloch (1898-1965) and Walter H. Pitts (1923-1969) in 1943, was pronounced by Marvin Minksy (1927-2016) and Seymour Papert (1928-2016) in their 1969 book \emph{Perceptrons: An Introduction to Computational Geometry} (note the subtitle)\cite{Minsky69}.

Rosenblatt's early success demonstrated that what is today called an artificial neural net may, but need not, allow a conceptual interpretation of its hidden units (the intermediate nodes between input and output). In the Gestalt-Holist view, nodes or pattern of nodes are not required to actually pick out fixed (micro) feature of the domain. Minsky and Papert's destruction of the original Perceptron is as much based on the technical limitations of the one-layer contraption in failing to calculate mathematical (symbolic) functions such as parity, connectedness and logical ``exclusive or (XOR)'' as on the philosophical defense of a  rationalist tradition supporting representational states of mind and symbolic manipulation against holism.

\subsection{The Advent of Modern Computational Minds: Connectionism and Parallel Distributed Processing}
\label{sec:orgc39f02a}
The counter-arguments, inter alia, contributed to tip the internal war between the two research programs towards the surprising asymmetrical victory of the symbolic representation scientific community. Those researchers who proposed using the (modern) digital computer as symbol manipulators in the 70s and early 80s had almost undisputed control of the research funding resources, graduate programs, research journals, symposia and conferences and started making inroads into connected communities in psychology, operations research, management and philosophy.

Nonetheless small research groups around David E. Rumelhart (1942-2011), James L. McClelland of the Parallel Distributed Processing (PDP) group , Geoffrey E. Hinton, Terrence J. Sejnowski, John R. Anderson, often sourcing their model scenarios from psycholinguistics and memory studies, looked to \emph{parallelism} as a theoretical and practical response to the previous decade attacks on connectionism. So quick and striking were their successes in the early and mid-eighties that Paul Smolensky \cite{Smolensky1988} could write in 1988:
\begin{quote}
``In the past half-decade the connectionist approach to cognitive modeling has grown from an obscure cult claiming a few true believers to a movement so vigorous that recent meetings of the Cognitive Science Society have begun to look like connectionist pep rallies.''
\end{quote}
Artifacts produced in this program, and variously called \emph{Boltzmann Machines}, \emph{Hopfield Nets}, \emph{PARSNIP}, \emph{Parallel Distributed Processing Architecture)}, and their surprisingly efficient performances soon found their way to wider public exposure both in the scientific and general press of the time.
 In 1987, for instance, a review of the PDP Group's published works in the \emph{New York Times} (soberly) concluded:
\begin{quote}
``[\ldots{}] connectionism's recent development as a promising line of research, and if future work is successful the results will constitute a major scientific achievement. The effort to explain the structures of information in human cognition, rather than merely to analyze and describe them, should be a major activity of cognitive science, especially cognitive psychology. That effort has been less salient lately than it deserves to be, and the promise of its revitalization is important - in the field and for its contribution to scientific knowledge at large.''
\end{quote}

\subsection{The Advent of Modern Computational Minds: The Physical Symbol System Hypothesis}
\label{sec:org4837341}
While connectionism struggled only to reappear in the late eighties with new models, Newell and Simon's GPS seeded the symbolic approach in a number of related research domains: computers, seen as systems for manipulating mental symbols, for instantiating representations of the world, for using mathematical logic to perform problem-solving as a paradigmatic model of thinking. This understanding of computer models became the basis of a way of understanding minds, as captured by \emph{The Physical Symbol System Hypothesis} \cite{Newell1982} stated by Newell and Simon:
\begin{quote}
``A physical symbol system has the necessary and sufficient means for general intelligent action.''
\end{quote}

So intense was this philosophical conviction than, even though in the meantime the exaggerated claims of the AI symbolists were themselves questioned in the famous review of academic AI research compiled  by Sir Michael James Lighthill (1924-1998)--which started what is sometimes referred to as the ``AI Winter''--the symbolic approach to general intelligent action was still considered more successful than connectionism. The way it looked then was that connectionist systems confronted a tremendous amount of mathematical analysis and calculating to solve even the most simple problem of pattern recognition. Symbol system, in contrast, blazed through problems like theorem proving and hard puzzles in logic relatively effortlessly. At the time, given the speculative nature of neurosciences, the computing power generally available, and the first commercial successes of practical AI applications \cite{Klahr1986,HayesRoth1983,McDermott1982} programs based on symbolic representations and logic processing were more than any others on their way to being of general use.
\subsection{The Philosophical Critique  (1988)}
\label{sec:org62435f6}
As had happened thirty years before, the sometimes extravagant claims of both the now 30-year established symbolic school and the reinvigorated PDP connectionist school of thought drew intense critical fire from philosophical quarters. Maybe two of the most devastating attacks came from Hubert Dreyfus again, on the one hand, in his paper co-authored with his brother Stuart \emph{Making a Mind vs. Modeling the Brain: AI Back at the Branchpoint} \cite{Dreyfuses1988}, and from Jerry A. Fodor (1935-2017) and Zenon W. Pylyshyn, on the other hand, in the well-known book chapter \emph{Connectionism and Cognitive Architecture: A Critical Analysis} \cite{Fodor1988}, both published in 1988.

The Dreyfus (family) paper aimed at defending the holistic inspiration of PDP connectionism against the anti-gestalt prejudice they saw in the symbolic approach. But they also criticized the symbolists, focusing on the vexing everyday knowledge or common-sense problem that plagues symbol systems at scale.

From a different viewpoint, Fodor and Pylyshyn engaged a broad clarification of the ``variety of confusions and irrelevances'' that marred discussions of the relative merits of the competing architectures of Connectionist Networks and ``Classical Computer'' models. The latter they consider, in 1988, were traditionally established in Cognitive Science. The authors stated discussions of cognitive architecture are exclusively about the representational states and processes, at the level of explanations of the mind qualifying as cognitive. From this standpoint, Fodor and Pylyshyn viewed both connectionists and classicists firmly in the Representationalist camp (as contrasted with the Eliminativists, who think that a theory of cognition could do without semantic representation, and stay with the neurological or possibly behavioral level). They proceeded to investigate whether the PDP connectionism could provide this level of explanations of the mind, and based on empirical findings of psycholinguistics, concluded negatively.

In their final remarks, Fodor and Pylyshyn saw the only worthwhile role of connectionism as one of mere implementation of the classical symbol system model, and suggested giving up the idea that networks, quoting Rumelhart and McClelland, were ``a reasonable basis for modeling cognitive processes in general'', and dropping the notion that all learning consists of a kind of statistical inference realized by adjusting parameters.

\subsection{The Philosophical Critique (beginning 2018)}
\label{sec:org6b42cd0}
Dramatic successes in Machine Learning in the recent contemporary period, have led to increasing expectations, in academia and in the larger public, for autonomous systems that exhibit human-level intelligence. These expectations, however, have met with what a third round of critique in the 30-year cycle deems fundamental obstacles.
\subsubsection{Crouching Symbols, Hidden Layers}
\label{sec:org3b13a15}
As already put forward by defenders of the 1988-vintage of PDP connectionism, today's philosophical critique of the third wave of connectionism would be ill-advised to saddle the ``new new connectionism'' with the sins and omissions of previous theories. In 1988, Paul Smolensky published a rebuttal of the arguments developed in the Fodor and Pylyshyn paper, calling \emph{On the Proper Treatment of Connectionism} (PTC) \cite{Smolensky1987,Smolensky1988}. PTC informed the whole next-generation neural net research work which, in turn, lead to the unquestionable successes of recent years.

\paragraph{The Proper Treatment of Connectionism}
\label{sec:org4f66e10}
Smolensky considered what role a connectionist approach might play in cognitive modeling, as a topic of cognitive science. He  traced a first source of confusion to the conflation of public knowledge and personal knowledge. While it is sound for scientific purposes of definition, communication and transmission, to formalize knowledge in linguistically expressed symbols and rules--what Smolensky calls \emph{cultural knowledge}--it is to be contrasted with \emph{individual knowledge}, skills, and intuition which are not driven by the same purposes, and might not be expressed linguistically.

Smolensky consequently posited two ``virtual machines'' in and of the mind, a division reminiscent of the S1/S2 systems of Daniel Kahneman \cite{ref9910114919702121}: one for conscious application of language-like rules and the other for ``intuitive processing''. It is then shown that sequential processing of linguistically formalized rules, championed in the symbolic approach, had to be rejected as a mechanism appropriate for the intuitive processor in favor of a properly stated connectionist approach. PTC rejects that elements in the intuitive processor refer essentially to the same concepts as the ones used to consciously conceptualize a task domain in the symbolic processor.

This position statement clearly contradicts Fodor and Pylyshyn view of the universality of the symbolic processing, the language of thought hypothesis, and their somewhat derisive suggestion that the sole worth of neural nets could be in possibly implementing that symbol architecture. Instead Smolensky argues for an autonomous subsymbolic level, with ``intuitive'' knowledge, distributed in the hidden layers, enriching the higher-level conceptual level and not of replacing it. The \emph{Subsymbolic program}, still very much active today, aims at explaining the successes and failures of the physical symbol systems, and at  accounting for the emergence of symbolic computation out of subsymbolic computation. 

\paragraph{Common-Sense: Scope and Limits of Symbolic AI}
\label{sec:orgda23fa9}
As for Fodor and Pylyshyn's recommendation to confine connectionist architectures to a yet elusive implementation level for higher models, Hector J. Levesque in a paper published November 1988, in the \emph{Journal of Philosophical Logic} \cite{Levesque1989}, sought to balance it with the computational burden of using logic to implement the complexity of reasoning. Given its apparent computational difficulty, it seemed to the author quite unlikely that logic could be at the root of normal, everyday thinking. This resembles the common-sense argument viewed from a computational resources perspective: how is it, for instance, that we can effortlessly answer questions like ``Could a crocodile run the steeple-chase?'', when we obviously could not and do not simply remember, or perceive, their answers?

Thirty years later, Levesque \cite{Levesque2014} pursues and revises this critical appraisal, with the hindsight of three more decades, in the same tone of voice we found in the first Dreyfus paper in the early sixties. Why he asks, after sixty years, have we still so little to show for knowledge representation regarding the science of intelligent behavior? The ``Knowledge is Power'' slogan of the AI classicists in the late eighties seems quite too facile today. The hurdles to symbolic AI still stand on the start of a third cycle:
\begin{quote}
\begin{itemize}
\item ``Much of what we come to know about the world and people around us is due to our use of language. And yet, it appears that we need to use extensive knowledge to make good sense of this language.''
\item ``Even the most basic child-level knowledge seems to call upon a wide range of logical constructs. And yet, symbolic reasoning over these constructs seems to be much too demanding computationally.''
\end{itemize}
\end{quote}

\paragraph{All The Turing Tests Couldn't Put Symbols Together Again}
\label{sec:org2178e87}
Note the parallax in Levesque's criticism addressed to the symbolic AI research program: where Smolensky was interested in confronting an emerging cognitive science to AI research--in fact, to its prominent research agenda at the time--Levesque is interested in intelligent behavior and its confrontation to ``the computational story'' told by the symbolic AI research program.  (Although it could similarly be addressed to today's connectionists). And this parallax inevitably harks back to the famous Turing Test. 

In the AI literature, the Turing Test for intelligent behavior is more often than not over-simplified \cite{Collins1990} into a question-answering game, played in writing--so that tones of voice or accents are irrelevant-- between participant \emph{A} and human ``judge'' \emph{J}, at the end of which judge \emph{J} tells whether participant \emph{A} is human or machine--an AI program, a software object, in the modern variant. The machine playing \emph{A} is successful at exhibiting intelligent behavior if it tricks judge \emph{J} into thinking that \emph{A} is human. Levesque's reproach to the Test is that a bag of really cheap tricks might fool \emph{J} but would certainly not qualify as general intelligent behavior-- wouldn't it fall clearly into Smolensky's cultural knowledge, and thus be relevant to critique of the symbolic school of AI only?

However that over-simplification is completely missing the subtlety in the original \emph{Imitation Game} proposed by Turing \cite{Turing1950}:
\begin{quote}
``The new form of the problem can be described in terms of a game which we call the `imitation game.' It is played with three people, a man (A), a woman (B), and an interrogator (C) who may be of either sex. The interrogator stays in a room apart front the other two. The object of the game for the interrogator is to determine which of the other two is the man and which is the woman. He knows them by labels X and Y, and at the end of the game he says either `X is A and Y is B' or `X is B and Y is A.'

We now ask the question, 'What will happen when a machine takes the part of A in this game?' Will the interrogator decide wrongly as often when the game is played like this as he does when the game is played between a man and a woman? These questions replace our original, 'Can machines think?' ``
\end{quote}
The game, in the original, is way more challenging: in order to win, the machine \emph{M} has now to convince judge \emph{C}, man or woman, that it is a woman \emph{B} pretending to be a man \emph{A}. The double imitation difficulty of the original game would make it quite unlikely that the proverbial bag of cheap tricks, that so irritates Levesque, proved sufficient for machine \emph{M} to win repeatedly.

Levesque suggests replacing the (over-simplified) Turing Test with what he termed \emph{Winograd Schema} questions such as:
\begin{quote}
\begin{itemize}
\item ``Joan made sure to thank Susan for all the help she had given. Who had given the help, Joan or Susan?''
\item ``Joan made sure to thank Susan for all the help she had received. Who had received the help, Joan or Susan?''
\end{itemize}
\end{quote}
which mix systematicity and compositionality, in Fodor and Pylyshyn's acception, in finding the referent of the pronoun in the sentences. Common-sense knowledge would be of course involved in answering, as in this other example:
\begin{quote}
\begin{itemize}
\item ``The trophy would not fit in the brown suitcase because it was so small. What was so small, the trophy or the suitcase?''
\item ``The trophy would not fit in the brown suitcase because it was so big. What was so big, the trophy or the suitcase?''
\end{itemize}
\end{quote}
Consideration of the original imitation game also suggests other variants for assessing intelligent behavior (indeed pushing further the cause of Levesque's irritation): the \emph{Deception Game} which consists of twenty-seven rounds of Turing's original Imitation Game, exercising all permutations of woman \emph{B}, man \emph{A} and machine \emph{M} in the roles of participants \emph{X} and \emph{Y}, and judge \emph{C}; followed by comparison of scores on the decisions of \emph{C}.

\paragraph{Scope and Limits of The Proper Treatment}
\label{sec:org0973388}
Hidden layers are also the nemesis in the contemporary criticisms directed at the interpretability of the current crop of connectionist models, as they used to be in the previous round. Indeed as Machine Learning systems penetrate critical society areas such as medicine, the criminal justice system, and financial markets,  an increasing number of observer point that the inability of humans to explain neural nets' results becomes acutely problematic.

Zack Lipton's illuminating discussion \cite{Lipton2016} of what desiderata are to be satisfied by a proper notion of interpretability shows the notion itself to be part of the problem. Firstly, Lipton notes, ``The demand for interpretability arises when there is a mismatch between the formal objectives of supervised learning (test set predictive performance) and the real world costs in a deployment setting.'' Such desiderata include:
\begin{quote}
\begin{itemize}
\item Trust
\item Transferability
\item Causality
\item Informativeness
\item Fair and ethical decision-making
\end{itemize}
\end{quote}
and hence the expected properties of interpretable models fall into two broad categories: \emph{how does the model work} (transparency) and \emph{what does the model tell me} (post-hoc explanations). 

The author's discussion of state-of-the-art algorithms in Machine Learning yields to the following interim takeaways in the current round of critique:
\begin{quote}
\begin{itemize}
\item Linear statistical models, often resorted to for approximately explaining how deep models work, are not strictly more interpretable than deep neural networks.
\item Claims about interpretability should be qualified, as there are many definitions of interpretability out of the non exhaustive list of desiderata above.
\item In some cases transparency maybe at odds with broader objective of Machine Learning or AI. Lipton indicates: ``As a concrete example, the short-term  goal  of  building  trust  with  doctors  by  developing transparent models might clash with the longer-term goal of improving health care. We should be careful when giving up predictive power, that the desire for transparency is justified and isn't simply a concession to institutional biases against new methods.''
\item Post-hoc interpretations can potentially be misleading. Lipton cautions: ``against blindly embracing post-hoc notions of interpretability, especially when optimized to placate subjective demands. In such cases, one might--deliberately or not--optimize an algorithm to present misleading but plausible explanations.''
\end{itemize}
\end{quote}
\paragraph{Is It Symbols All The Way Down?}
\label{sec:orgd53553d}
Worries about solving the right problem with models are not only Lipton's in the current round of critique in the 30-year cycle. The interpretability obstacle is in fact inherent to both connectionist and symbolic approaches, whether old or new. The \emph{symbol grounding problem} had been identified by Stevan Harnad \cite{HARNAD1990335}, as early as in our second cycle in the discussion on AI: ``How can the semantic interpretation of a formal symbol system be made intrinsic to the system?'' In Charles Travis' words \cite{Travis2001}, referring to Putnam, ``how are mind and world dependent on each other?'' How can the meaning of the arbitrary, according to the physical symbol hypothesis, tokens, manipulated solely on the basis of their conventional shape, be grounded in anything but other meaningless symbols? If not, its semantic interpretation is nothing less than parasitic on the meaning on our heads and we are back to issues of representation raised by Putnam and by Kripke.

Harnad's solution is to hybridize symbols and neural nets with connectionism a natural candidate for the missing mechanism that learns the invariant features underlying categorical representations in reasoning. Connectionism links names to the proximal projections of the distal objects they stand for. Symbolic functions emerge as an intrinsic, dedicated symbol system relying on a bottom-up grounding of category names in their sensory representations.

This solution, however, leaves intact the criticisms of the Dreyfuses as of the relevance of the learned invariant features in the larger context of transferring to a different task for instance. Moreover, lessons learned from adversarial learning in the current generation of Deep Learning nets draw attention to how volatile the learned features could be, blurring even further the issue of interpretability.

\subsubsection{Deep Learning Without A Cause}
\label{sec:org6f48ef1}
Reminiscent of Dreyfus' survey at the first cycle of the discussion, Gary Marcus \cite{Marcus2018} reviews the state-of-the-art results of the widely publicized Deep Learning variety of connectionism:
\begin{quote}
``Before the year (2012) was out, deep learning made the front page of \emph{The New York Times}, and it rapidly became the best known technique in artificial intelligence, by a wide margin. If the general idea of training neural networks with multiple layers was not new, it was, in part because of increases in computational power and data, the first time that deep learning truly became practical.''
\end{quote}
The pattern Dreyfus identified two 30-year cycles ago can be readily observed again:
\begin{quote}
``A recent New York Times Sunday Magazine article, largely about deep learning, implied that the technique is `poised to reinvent computing itself.''' \cite{NYT2016}
\end{quote}
However, Marcus warns, the proverbial wall is near. Ten challenges are identified as critical:
\begin{quote}
\begin{itemize}
\item Deep Learning thus far is data hungry, in contrast humans can learn abstractions in a few trials. This is revisiting the issues of which micro-features are abstracted out in the hidden layers.
\item Deep Learning thus far is shallow and has limited capacity for transfer. Confronting a deep learning neural net to scenarios differing, even slightly, from the ones it was trained on shows that learning was superficial.
\item Deep learning thus far has no natural way to deal with hierarchical structure. This signals back to the language of thought hypothesis in discussions at the second round of the 30-year cycle.
\item Deep learning thus far has struggled with open-ended inference. This line of discussion expands on the Winograd schemes proposed by Levesque as appropriate tests of intelligent behaviors.
\item Deep learning thus far is not sufficiently transparent. The discussion about the opacity of ``black-box'' neural networks ramifies from the psycholinguistic angle in the previous rounds of critique to the current explainability or interpretability obstacle mentioned in the previous section.
\item Deep learning thus far has not been well integrated with prior knowledge. Marcus notes that  ``researchers in deep learning appear to have a very strong bias against including prior knowledge even when (as in the case of physics) that prior knowledge is well known. It also not straightforward in general how to integrate prior knowledge into a deep learning system, in part because the knowledge represented in deep learning systems pertains mainly to (largely opaque) correlations between features, rather than to abstractions like quantified statements (e.g. all men are mortal), or generics (violable statements like dogs have four legs or mosquitos carry West Nile virus).''
\item Deep learning thus far cannot inherently distinguish causation from correlation.
\item Deep learning presumes a largely stable world, in ways that may be problematic.
\item Correlatively,  deep learning thus far works well as an approximation, but its answers often cannot be fully trusted. A growing array of papers focus on how deep learning nets can be easily fooled, showing vulnerability to cheap, and sometimes not so cheap like in adversarial learning, Levesque tricks.
\item Deep learning thus far is difficult to engineer with.
\end{itemize}
\end{quote}

Whether these challenges, taken collectively, mount a critical attack against the third generation connectionism--and the author's cautious approach can be read in the ``so far'' phrase--and warrant new responses from the neo-neo-connectionists, beyond last-cycle-old PTC, is yet to be seen. A novelty, in the list of challenges, is however apparent when compared to previous critical arguments. To the obstacles of adaptability and interpretability, the previous round of discussion already honed in on, explicit understanding of cause-effect relations is added. (Although it was argued back then that rules in the symbolic approach could be used to capture causal inferences as well as other kinds of inferences. Thus this line of discussion really belongs in the perennial symbolism-connectionism confrontation.)

Today's form of the causality argument, to quote Judea Pearl \cite{Pearl2018}, ``postulates that all three obstacles mentioned above require equipping machines with causal modeling tools, in particular, causal diagrams and their associated logic.'' From a different standpoint, this says that theoretical limitations of model-blind machine learning, such as Deep Learning, would not apply to most tasks of prediction and diagnosis, with the provision of addressing the remaining challenges in Marcus' list though, but only to domains where reasoning about interventions (\emph{What if?}) or counterfactuals (by imagination or retrospection) proves essential, e.g. scientific thinking as well as legal and moral reasoning. While underlying notions of industrial applications and impact in economics and society are obviously at stake in this rephrasing of the argument, it points again to hybrid architectural solutions as in the previous round of the 30-year cycle.

\subsubsection{Neuromorphic Systems Reinvented}
\label{sec:org882e67c}
The argument of \emph{neural plausibility} is, interestingly enough, absent, or at least subdued, in the current round of critique. Early philosophers and neurologists interested in cognition, like Herbert Spencer (1820-1903, John Hughlings Jackson (1835-1911) and Charles Sherington (1857-1952) had sophisticated views of brain processing, based on the position that the central nervous system was a complex product of a long evolution. 

Since McCulloch and Pitts, and Rosenblatt's Perceptron, connectionism has been, most of the time in the past cycles, physiologically extremely implausible at the neuronal level--and sometimes deliberately so. Models proposed by McCulloch and Pitts, Hebb, Hopfield, Rosenblatt, Smolensky and others were avowedly simplified in order to capture only the essential features of the models they implemented, and simply in view of the computing resources required for their simulation.

Several workers, however, kept the neural plausibility concern alive in their research across the returning cycles of the AI claim-critique pattern. Terrence Sejnowski \cite{Sejnowski1976,Sejnowski1988,Sejnowski2018}, Patricia Churchland \cite{Churchland1988,Churchland1992}, Michael Arbib \cite{Arbib1972} and Stephen Grossberg \cite{Carpenter1987,Grossberg1988}, to name only a few, directly modeled their neural networks from consideration of the physiology of the brain. (And particularly in investigating all non-linearities in unit processing of the elementary nodes in neural networks.)

Even though much less publicized than today's Deep Learning ``revolution'', which claims prompted the current round of critique, this line of work also benefited from the rapid progresses in computing power of the past three decades. Large-scale neural (brain) simulations are becoming increasingly common \cite{deGaris2010,Goertzel2010}. (One billion neurons simulations were already reported in 2007 \cite{Ananthanarayanan2007}.) 

Current state-of-the-art neuromorphic systems offer a distinctly functional view and set of hypotheses regarding the neural mechanisms and organization that may underlie basic cognitive functions \cite{Eliasmith2012}. Arguments  contrasting the neuromorphic approach with connectionism or with the symbolic approach are not yet collected into a coordinated critique of AI. Whether  neuromorphic systems then will substitute for Deep Learning and recurrent neural networks in a yet to come fourth-round of the 30-year cycle remains to be seen.
\section{Structure}
\label{sec:org3972d0a}
The AI research community generally offered their concrete realizations, often designated by suggestive surnames (\emph{Eliza}, \emph{Pandemonium}, \emph{Perceptron}, \emph{Logic Theorist}, \emph{General Problem Solver}, \emph{Boltzmann Machine}, \emph{AlphaZero}, etc.) as technical objects the very existence of which was intended to serve as a proof of the conceptual hypotheses underlying the mind-model they illustrated. While the first-cycle objects included electronic devices, similar to the cybernetics artifacts of the late 40s and 50s, all others, even early ones, were and are software artifacts, often operating at the peak of the computing power available at their times of inception.

From a methodological perspective, AI research in the 30-year cycle could be viewed as operating in the formalism of Mathematical Logic, with its artifacts playing the role of proof-carrying theorems in a larger theory of cognition. In doing so, however, it should be no surprise that these objects are, and were, interpreted quite differently by workers in other related disciplines such as linguistics, psychology, neurosciences, philosophy, which by the way, very much like AI itself, were simultaneously in flux. 

These ``boundary objects'', with sudden bursts of proliferation, and subsequent mass extinctions, appear then to return in about 30-year cycles, culminating respectively circa 1958, 1988 and 2018. At each peak, their genesis and construction almost completely determined the nature of the sometimes heated debates about their significance for the understanding of cognition, in the individual and sometimes in the collective.

\subsection{An Analytical Philosophy Background}
\label{sec:orgdb0b2d5}
The start of the 30-year cycle should first be cast against the historical perspective of a reaction to associationism. The psychology theory prevailed in the early twentieth century, as expressed in the work of the Wurzburg school. It introduced \emph{tasks}, as necessary factors in describing the process of thinking. It also relied on systematic introspection to shed light on the state of consciousness while performing tasks. Opponents to the latter developed a couple of divergent antitheses, Behaviorism and the Gestalt-movement, which both claimed the futility of introspection and asserted the mechanistic assumption of stimulus-response connections as simple determinate functions of the environment for the gestaltlists. Both rejected the mechanistic nature of associationism but reformulated the task-driven nature of thinking in terms of holistic principles in the organization of thought.

In fact the development of the modern computer from the mid-40s strengthened the associationist assumption in human perception and cognition initiated by 18th century philosophers David Hume (1711-1776) and George Berkeley (1685-1753). Associationism was expressed for instance in the stimulus-response psychology of 19th century Herbert Spencer (1820-1903), whose emphasis on physiological detail in biological evolution, however inaccurate, strongly influenced both psychologists and physiologists: thinking should be analyzable into simple determinate operations (in complete opposition to the gestaltists).

This is readily visible in the first-cycle discussions. Notwithstanding Newell, Shaw and Simon's 1958 claims to have synthesized a common view from associationist-gestaltist extremes, Dreyfus considered that strengthening as misconstrued when:
\begin{itemize}
\item empirical evidence of the associationist assumption can be questioned on the grounds of a critique of the scientific methodology in cognitive simulation, and
\item the apriori argument for associationism when properly rephrased turns inconclusive, ``we need not conclude from the claim that all physical processes can be described by in a mathematical formalism which can in turn be manipulated by a digital computer that all human information processing can be carried out as discrete processes on a digital machine''.
\end{itemize}
The criticism was all the more to the point, at the time, than the early 20th century thread of associationism in behaviorism, championed by e.g. John B. Watson (1878-1958) and Clark L. Hull (1884-1952), came to be generally abandoned or deeply transformed by contemporary and later behaviorists such as B.F. Skinner (1904-1990) and Edward C. Tolman (1856-1959).

It is in fact the structure of the cyclic discussion, poised against the background of the increasing prevalence of the man-machine metaphor in the analytical philosophy of mind, that helped define cognitive science as a distinct endeavor. In other words, the major question addressed by outside research communities in their analyses of the technical objects of AI, has been aptly expressed by Van Gelder in 1995 \cite{VanGelder1995} : ``What might cognition be if not computation?''

\subsection{From Mind-Body to Man-Machine}
\label{sec:org8d8caf6}
Turning then to analytical philosophy, the structure of the discussion is explicitly clarified by \emph{Minds and Machines} (1960) \cite{Putnam1960}, where Putnam revisited the centuries-long mind-body debate:
\begin{quote}
``The various issues that make up the traditional mind-body problem are wholly linguistic and logical in character.[\ldots{}] all the issues arise in connection with any computing system capable of answering questions about its own structure and have thus nothing to do with the unique nature (if it is unique) of human subjective experience.''
\end{quote}

This is the core of Dreyfus' critical appraisal of AI at the first revolution of the 30-year cycle. Indeed, the general interest in the so-called ``electronic brains'' was used by analytical philosophers, on the one hand, to recast the historical debate dividing behaviorists from Cartesians; and by moralists and philosophers, on the other hand, to claim that higher forms of human behavior were beyond the powers of any machine (not only of then current state-of-the-art electronic computers).

Both parties however, noted Dreyfus, ``credulously assumed that highly intelligent artifacts had already been developed.'' Dreyfus drew attention to how much the AI boundary objects of 1958 were psychologically unconvincing and therefore did not warrant the significance they were so generously granted.

\begin{table}[htbp]
\caption[Information Processing]{\label{tab:org92bade6}
Controversial features in the first cycle critique}
\centering
\begin{tabular}{ll}
\textbf{Human Information Processing} & * Machine Information Processing*\\
\hline
Fringe of consciousness & Heuristically guided search\\
\hline
Essence/Accident Discrimination & Trial and Error\\
\hline
Ambiguity Tolerance & Exhaustive Enumeration\\
\hline
Continuous processes & Discrete/Determinate Associationism\\
\hline
\end{tabular}
\end{table}

The argument is about the distinctive features of human versus machine information processing, as summarized in table \hyperref[tab:org92bade6]{Information Processing}. The table is used to disqualify a strong form of associationism--in which the assertion that human information processing is explicable in discrete, step-wise, symbolic only terms, is claimed to be based on and to explain descriptions of human experience and behavior--is untenable.

Dreyfus recognized, however, that the realization that basing cognition only on a representational associationist model (namely, the physical symbol system hypothesis) is unfounded, leaves untouched the weaker claim that human intelligent behavior, not information processing, can be simulated using digital computers. Although invoking the computing power argument, in considering the computer state-of-the-art current at the time Dreyfus wrote, he could raise serious doubts about the computing and time resources that would be involved in implementing the simplest physical process in the brain. Today the computing argument is rarely, if even, mentioned as it is generally assumed that available processing resources surpass the brain's.

As research product artifacts, LT and GPS were intentionally designed by their authors to ``maximally'' confuse the attempt to accomplish with machines the same tasks that humans perform, and the attempt to simulate the processes humans actually use to accomplish these tasks. Nonetheless, from the origin of the cycles, there seems to be a conceptual divide between appreciations of the  scientific worth of machine artifacts, computers and electronic devices in 1958, software and algorithms in subsequent cycles, that simulate or reproduce human cognition, and artifacts that implement physiological brain functions, in view of exhibiting intelligent behavior. This debate, and in its cyclic 30-year return, is both purposely anti-behaviorist and centered on the individual task-performing mind.

\subsection{What science for cognitive science?}
\label{sec:orgdb2e0a8}
By 1988 and the second peak in the 30-year cycle, the evolution of foundational ideas in the research communities parties to the AI discussion called for a reinterpretation of the technical artifacts it produced and had produced.

By that time, the ``cognitive science'' research program had been installed and stood on firmer grounds. In the same intervening three decades, the mounting prevalence of the discrete digital, ``associationistic'', computer as instrument and metaphor of the mind went almost unchallenged. The physical symbol hypothesis had become the orthodox view in AI research, with a minority of  research groups still working on alternatives, in the cold of this first AI Winter.

Meanwhile, technical progresses in computing processing, memory, algorithms and software narrowed the AI discussion to the confrontation of the new wave of PDP connectionism to the then well-entrenched orthodoxy of symbolic systems.

In this context, the Dreyfuses 1988 critique is still focused on the divide of the earlier cycle, inspired by their earlier anti-associationism campaign, although now confronting the then new Hinton- and Rumelhart-McClelland PDP brand of neural networks \cite{rumelhart86a} to the background context of representational cognitive systems, derived from the physical symbol system hypothesis. (Expert Systems had started to pop up in successful commercial applications \cite{Durkin1996}). Much of the philosophical structure in their 1958 criticism still held in 1988, although now informed (i) by their revisiting of Heidegger's work, into which Dreyfus had embarked in the meantime, and (ii) by the acknowledgment of the vexing problem of common-sense knowledge representation confronting symbolic systems. 

\begin{table}[htbp]
\caption{\label{tab:orgdb06d22}
The Mind-Brain Modeling Branchpoint in the 1988 AI critique}
\centering
\begin{tabular}{ll}
\textbf{Making a Mind} & \textbf{Modeling the Brain}\\
\hline
Symbolic information processing & Subsymbolic\\
\hline
Sequential discrete execution & Parallel distributed execution\\
\hline
Husserl, early Wittgenstein & Heidegger, late Wittgenstein\\
\hline
Common-Sense obstacle problem & Everyday know-how\\
\hline
\end{tabular}
\end{table}

The critique addressed both the symbolic school of AI and new connectionists. The latter, according to the authors, ``influenced by symbol manipulating AI, are expending considerable effort, once their nets have been trained to perform a task, trying to find the features represented by individual nodes or sets of nodes. But sometimes most nodes in neural nets cannot be interpreted semantically at all, so that effort is wasted and in fact misconstrued. Of course, any successfully trained multi-layer network has, if in a limited sense, an interpretation in terms of (distributed) features or symbols. The question is whether or not this interpretation preserves the rationalist intuition that these symbols must capture the supposedly invariant, essential structure of the task domain, and not some accidental features of the training dataset, to base a theory of the mind on them.'' 

Evidently this is a profound criticism, that resurfaces in today's episode on the interpretability of hidden layers, if differently expressed. Should it be even possible to find some meaning in the hidden layers of neural nets--and the Dreyfuses insist that this was questionable indeed and still is--how do we make sure that these hidden layers symbols are not just spurious side-effects of the nature or volume of the training set? In this critical view, 1988-vintage new-connectionism, and likewise 2018-vintage one may add, sheds no light on human cognition.

Furthermore, common-sense knowledge, argued as distinctive of human cognition  and falling clearly into the symbolic camp according to Smolensky \cite{Smolensky1988}, was similarly problematic for both parties \cite{Lenat1986,Lenat1988} at the time.

Both strong arguments were used in a critique, not of misconstruing human information processing as in the Dreyfus' review of early AI efforts, but of what would be the proper scientific method for a budding cognitive science. Shouldn't the ``cognitive science'' variant of AI research produce models that enlighten the understanding of human cognitions?

Compared to the previous round in the AI discussion, the novelty was then in shifting the target of the critique to the topic of scientific methods in cognitive science. The shift defines the background of the whole 1988 debate. For instance: in examining the vexing question he himself asked, Van Gelder \cite{VanGelder1995} offered that an alternative view to cognitive systems being computational could be that they simply be dynamical systems. (Ironically enough, in an operations-research sense which was much in currency circa 1958!) His discussion of the classical ``cybernetic'' task accomplished by the Watt centrifugal governor suggested a few principles to retain for a young cognitive science:
\begin{quote}
\begin{itemize}
\item various kinds of systems, fundamentally different in nature and requiring very different tools for their understanding, can subserve sophisticated tasks--including interacting with a changing environment;
\item our sense that a specific cognitive task must be subserved by a computational system may be due to deceptively compelling preconceptions about how systems solving complex tasks must work;
\item cognition may be the behavior of some (non computational) dynamical system.
\end{itemize}
\end{quote}
Picking up examples from Economics and Psychology, Van Gelder points to \emph{Decision Field Theory} \cite{Busemeyer1993} as illustrative of the latter point. The principles, however, leave open the question of whether or not the brain is such a dynamical system or a computational system.

In his proposed three-species classification of cognitive systems, new connectionism appears only as a hybrid:

\begin{table}[htbp]
\caption{\label{tab:org60a1ae6}
State-dependent Systems Classification by Van Gelder (1995)}
\centering
\begin{tabular}{ll}
\textbf{Species} & \textbf{Characteristics}\\
\hline
Computational & Classical Turing Machine model\\
\hline
Dynamical & States are numerical, rules of evolutions specify\\
 & sequences, often continuous (ODE), of such states.\\
\hline
Connectionism & They are dynamical systems in the above definition,\\
 & albeit specific: high-dimensional, homogeneous and\\
 & ``neural'' in the sense of local activation being a form\\
 & of integration of connected neighbors' activation.\\
 & But the connectionism of circa 1988 is found to\\
 & retain much of the basic structure of the computational\\
 & picture, maybe simply changing the nature of the\\
 & representations (what the symbols stand for).\\
\hline
\end{tabular}
\end{table}

\subsubsection{Internal Debate: Crossing The Divide}
\label{sec:org52db8b5}
The frontline, all within cognitive science itself, pitched defenders of the symbolic approach against neo-connectionists. The structure was set by the heated dialogue between Fodor and Pylyshyn, on the one hand, and Smolensky and followers, on the other. Heavy reference to analytical philosophy was present all over the debate. It was used to amplify
the vacillation, noted by van Gelder, of new connectionism between representationalism--the expensive effort to find a proper explanation of the performance of neural networks mentioned above--and the claim that the cognitive level is altogether disposable in favor of a biologically inspired one.

\begin{table}[htbp]
\caption{\label{tab:org7d0d22c}
Structure of the Insiders' 1988 Argument Against New Connectionism}
\centering
\begin{tabular}{ll}
\textbf{The Need for Symbol Systems} & \textbf{Failures of New Connectionism}\\
\hline
Productivity of thought: classical & Combination of parts as a basic\\
argument for the existence of & operation runs against the hypothesis\\
the existence of combinatorial & of parallel distributed processing.\\
structure in any rich enough & \\
representational system. & \\
\hline
Systematicity of cognitive & There might not be structural relations\\
representation. & between mental representations captured\\
 & by a neural net that correspond to\\
 & systematically linked thoughts.\\
\hline
Compositionality of representations, & Commitment to mental representations\\
which result as a consequence of & without combinatorial structure precludes\\
combining productivity and & compositionality. (Hard-liners are even\\
systematicity arguments. & tempted to reject it altogether.)\\
\hline
Systematicity of Inference: inferences & Neural nets are indifferent as to whether\\
of a similar logic type ought & similar nodes or sets of nodes represent\\
to elicit correspondingly similar & logical inferences of a similar type, as\\
cognitive capacities. & representations are distributed.\\
\hline
\end{tabular}
\end{table}

It is interesting to note that the crux of the dispute is articulated around features of language, such as productivity and systematicity, that had been well established by empirical psycholinguistics research in the intervening decades, and how they could be found or not in the neural net performance.

In the philosopher's eye, features of language are used as ``proxies'' for features of thought that a cognitive architecture should account for. This position is not really discussed in Smolensky's defense: \emph{Proper Treatment of Connectionism} (PTC) \cite{Smolensky1988,Smolensky1987}. In it a novel subsymbolic paradigm is suggested instead, as a solution to the richer behavior exhibited in psycholinguistic mind studies. This behavior appeared governed by complex systems of rules but nonetheless still awash in variance, flexibility and fluidity, deviation and exceptions. Posited between the neural level and the conceptual level, both of them unquestioned, this subsymbolic level would account for individual private knowledge, distinguished from a cultural and social form of  knowledge driven by conscious rule interpretation.

\begin{table}[htbp]
\caption{\label{tab:orgb8d8418}
Smolensky's Defense of the Subsymbolic Level}
\centering
\begin{tabular}{ll}
\textbf{The Simplistic (and Incorrect)} & \textbf{The PTC Defense}\\
\textbf{Analysis of Connectionism} & \\
\hline
All representations are atomistic. & Representing entire logical propositions\\
 & by individual nodes is not typical\\
 & of connectionist models.\\
\hline
Connectionist processing is strictly & Mental states are described by distributed\\
association, which is sensitive to & activation levels ascribing to them a\\
statistics, not to structure. & proper compositional structure, but in\\
 & an approximate sense, not equivalent\\
 & to algebraic combinations of\\
 & context-independent representations.\\
\hline
One class of knowledge & Distinct individual and socio-cultural\\
 & knowledge bodies and representations\\
\hline
\end{tabular}
\end{table}

PTC is used by Smolensky to rebuke Fodor and Pylyshyn's claims \cite{Smolensky1987}: 

\begin{itemize}
\item connectionist models do support complex representations found in symbolic systems based on composition of, in PTC distinctively, distributed, representations of constituents;
\item if the constituency relationship among distributed representations is important for the analysis of connectionist models--still with the back-of-the-mind idea to emerge symbols from activation patterns in the hidden layers--it is emphatically not a causal mechanism within the model; there is no necessity for a neural net, in order to process a complex representation, to decompose it into its constituents, a property already noted in the Dreyfuses analysis. The arbitrariness in which the constituents are defined in the hidden layers introduce no ambiguity in the way the representation is processed. The ambiguity is in the eye of the beholder: attempts to formalize the notion of constituency in PTC crucially involves approximation.
\end{itemize}

PTC, as a tentative rebuttal of Fodor and Pylyshyn's language-of-thought inspired critique, calls on the familiar reductionism to try crossing the divide between artificial neural nets and symbolic representations. Actually it conflates the claim that mental states may be reduced to, or analyzed into, physical states and the weaker claim that one class of states could not exist without the other. Hence the various forms of hybrid technical objects that were proposed to solve the conundrum \cite{Langley2017}\cite{Jilk2008}\cite{Smolensky1987}.But, in contrast to Frege's ``eternal thoughts'', the 1988 debate has been narrowed to discussing reduction of \emph{representations} in symbols and logic to \emph{models} of neuronal events in connectionist artificial nets, both simulated \emph{in silico} by software programs--so they both reduce to physical events after all, and hence are only a matter of language.

\subsubsection{External Debates: Foundational Challenges}
\label{sec:orgf7ed312}
While the Cognitive Science program matured as a valid, independent line of research, discussions also went unabated in the related disciplines from which most of the critique originated. Some of the questions raised in debates in psychology and in the philosophy of mind also impacted the previous and current interpretations in the AI critique. Arguments, explicitly evoked in the 1988 wave, proved not immune themselves to criticism from outside AI circles.

\paragraph{Questioning Systematicity}
\label{sec:org0878a5d}
In examining the systematicity of language and thought, an argument of Fodor and Pylyshyn's against PDP connectionism, Kent Johnson (1970-2017) expressed skepticism \cite{Johnson2004} about the fundamental claims of psycholinguistics that:
\begin{itemize}
\item systematicity of language is taken to be trivially true;
\item systematicity is so clearly defined that it needs only a cursorily sketch.
\end{itemize}

Johnson's analysis convincingly shows that natural intuitive characterizations of the claim that language is systematic are false and that systematicity of thought is at least as problematic as systematicity of language, \emph{pace} Fodor. In fact, the linguistic natural kinds with which language is working do form a complex overlapping structure rather than a clear-cut partition--the approximation defended by Smolensky above.

Given this complex structure, it becomes unclear what it would be for naturals language to be systematic, and the jump to systematicity of thought appears even more elusive. Does that vindicate new connectionism? Not really: the evidential status of connectionism is unaltered. Indeed it may well be thought that a connectionist artifact, properly trained, has indeed learned a systematic language, as Smolensky advocates in his PTC, but in doing so has only induced an empirically an incorrect organization of natural kinds of expressions, as the Dreyfuses suspected in their 1988 paper.

Systematicity of the lexicon has also been challenged: Jeffrey Elman (1948-2018) argued that empirical data are problematic for the view of the lexicon--the inventory of symbols in the representational model--as an enumerative database \cite{Elman2014}. (Which could count, by the way, as a minor failure of the computer-model of the mind \cite{Tulving2003}.) The richness of lexical knowledge, as demonstrated by this ``workhorse'' of the psycholinguistic literature, the study of how comprehenders--human or artificial--process sentences, is revealed in its quite detailed nature. The processing is  idiosyncratic and verb specific. Lexical knowledge is brought to bear at the earliest stage possible in incremental sentence processing but dynamic factors such as the verb's grammatical aspect, and the agent and instrument involved, may significantly alter this process.

The idea that the form of the lexicon has no consequences for symbol processing is an essential tenet of the physical symbol system hypothesis. But that entails that performance tells us little about competence and that conclusion defeats the purpose of studying cognitive architectures in silico as a scientific method. Elman suggested an alternative to the lexicon-as-dictionary with a lexicon as a dynamical system, a la Van Gelder, and implemented a model demonstration as a recurrent neural network, in which feedback loops allow the current state of the system to be affected by prior states. In doing so, Elman himself suggested an hybrid technical object for consideration as a cognitive architecture.

\paragraph{Questioning Semantic Categories}
\label{sec:org3bfb38b}
In her pioneering psychological study of cognitive representations, Eleanor Rosch \cite{Rosch1975} demonstrated that:
\begin{quote}
\begin{itemize}
\item ``the internal structure of superordinate semantic categories was clearly a pervasive aspect of the way in which they were processed in the tasks'' in the experiments; (cognitive representations appeared to be more similar to the good examples than to the poor examples)
\end{itemize}
\begin{itemize}
\item ``meaning could affect perception'', there are several levels of processing over which internal structures of categories have selective effects;
\item ``there is tentatively a meaning of superordinate categories, not specifically coded in terms of words or pictures but translated into a format in preparation for actual perception.''
\end{itemize}
\end{quote}

These findings supported the emergence of an ambitious theory, or more properly a class of theories, in the psychological research of the seventies and the eighties. Generally called \emph{Prototype Theory} it states that entities fall neither sharply in nor sharply out of a concept's extension. An object instantiates a concept to the extent that it is similar, in a properly defined measure, to the prototype of the concept. 

While Prototype Theory, in construing membership in a concept's extension as graded, would seem to call on psychological empirical research to help Smolensky's PTC claim of approximation as a crucial ingredient in a cognitive architecture, Daniel Osherson and Edward E. Smith (1940-2012) revisited its compatibility with two criteria of adequacy for full-fledged theories of concepts, namely the relationship between complex concepts and their constituents (combinatorial compositionality in the 1988 critique), and the truth conditions for thoughts corresponding to simple inclusions \cite{Osherson1981}. 

The truth condition characterizes the circumstances under which thoughts, taken to be immediately constituted by concepts, are true in classical syllogisms (\emph{All A's are B's}, \emph{Some B's are C's} and so forth). Considering many forms of measure of psychological similarity used in Prototype Theory--distance, number of common features, weighted contrast between common and distinctive features--the authors systematically derived logical contradictions in the truth values Prototype Theory assigned to composed thoughts.

Interestingly enough, the contradiction always stemmed from the possibility of two constituents (``crocodiles'' and ``steeple-chase runners'') of a logically empty set (``crocodiles that can run the steeple-chase'') to have individual non-null characteristic functions, and therefore strictly positive when combined. Also ambiguities appear in combined set definitions (``some crocodiles, say this particular alligator, could, that ran the hundred-meter hurdles'').

The reconciliation offered by the authors between traditional theories of concepts and prototype theories, namely that the former should be used for core concept procedures explicating its relation to other concepts while the latter should be used for identification procedure specifying the kind of information to make rapid decisions, did not seem to argue in favor of disposing of either one of the symbolic and subsymbolic levels. The proposed reconciliation is in fact not only reminiscent of the division between reasoning and perception which underlies most of the critique, but would favor another form of hybrid solutions.

With basic hypotheses of the symbolists' critique of connectionism being themselves questioned on logical and psychological grounds, the great divide in cognitive architecture seems even more challenging to cross.

\section{Conclusion}
\label{sec:org0a9ae4c}
It is apt and relevant, if ironical, at the end of spinning out three times the philosophical arguments in the recurring critique of AI, to quote Putnam's somewhat radical conclusion of his \emph{Mind and Machine} paper, back at the eve of the first cycle:
\begin{quote}
``the mind-body problem is strictly analogous to the problem of the relation between structural and logical states [in a Turing Machine], not that the two problems are identical. [\ldots{}] if the so-called mind-body problem is nothing but a different realization of the same set of logical and linguistic issues, then it must be empty and just as verbal.
\end{quote}
Consideration of the abundant literature making up the on-going discussion on the AI critique scene sheds light on an apparent 30-year recurrence of waves of philosophical criticisms peaking around 1958, 1988 and 2018. It starts with an observable pattern, ``early, dramatic success followed by sudden unexpected difficulties'', generating assaults from related scientific disciplines, mainly psycholinguistics, psychology, philosophy and increasingly--maybe even more so in further cycles if any--neuroscience. A form of ``AI Winter'' generally follows the latter heated exchanges, which invariably center around interpretations of the technical productions of AI--electronic devices in cybernetic times and in the early cycle, increasingly sophisticated software artifacts in later cycles--that beg to differ.

Often in the deep of these AI winters, muted theoretical advances are made in unabated research communities which, combined with the (definitely non-cyclic but rather linear, if not polynomial) advances in computing processing power and memory, feed he next generation of dramatic successes. Looking at these successive technical artifacts as boundary objects helps charting a taxonomy of the arguments in the subsequent exchanges of critique and defense.

In the tradition of analytical philosophy, Cognitive Science was born in a new attempt at reductionism based on the advent of the modern computer, trying to cross the brain-mind divide using machine software both as a metaphor and an instrument. In a 30-year cyclic repetition of a pattern identified early on by Dreyfus, the critique of the technical successes in AI shows invariant structure: postulating intermediate levels of explanations, usually one, called ``subsymbolic'',  to help cross the divide or to show reduction, or abstraction, to be altogether impossible. Additional pro and contra arguments, developed in outside disciplines related to AI, were opportunistically brought in at different times. Today's neuromorphic systems could be said to add one more hypothetical biologically-inspired level into the mix, adding more philosophical questions about analyzing any level into others.

Each return of the cycle showed a shift in the core debate. Circa 1958, the question of comparing information processing in humans and in machine was critical. Circa 1988, the debate around competing cognitive architectures, and their possible hybridization, relayed the underlying concern about the proper scientific methods and goals of Cognitive Science. But in both occasions, most parties to the discussion credulously assumed that highly intelligent artifacts had already been developed. The reiterated difficulties in addressing the challenges of a maturing cognitive science repeatedly show how premature this hypothesis was.

And still is. The shift of the current critical debate on AI is increasingly towards ethical and socio-economics questions. It is observable in a time when concerns about AI \cite{National2018} are escalated to national levels, and the new dramatic successes of technical artifacts (autonomous cars, lethal autonomous weapons, drug discovery and personalized medicine, smart courtroom machines, board game and video game players, voice-controlled home systems, etc.) are the current boundary objects. Dystopian predictions about \emph{General Artificial Intelligence}, \emph{Singularity}, or \emph{Superintelligence} have prompted administrations and governments to start ``AI Governance'' multidisciplinary research programs. But Dreyfus's admonition of 1958 on the prematureness of interpretation still rings true, and  the sobering conclusions of Putnam should indeed not be forgotten:

\begin{quote}
``To put it differently, if the mind-body problem is identified with any problem of more than conceptual interest (e.g. with the question of whether or not human beings have `souls'), then \emph{either} it must be that (a) no argument \emph{ever} used by a philosopher sheds the \emph{slightest} light on it, or (b) that some philosophical argument for mechanism is correct, or (c) that some dualistic argument does show that \emph{both} human beings \emph{and} Turing Machines have souls! I leave it to the reader to decide which of the three alternatives is at all plausible.''
\end{quote}

\bibliography{cycle30y}
\end{document}